# NLP and CALL: integration is working

Georges Antoniadis[1], Sylviane Granger[2], Olivier Kraif[1], Claude Ponton[1], Virginie Zampa[1]

[1] Université Stendhal de Grenoble, Laboratoire LIDILEM
{Georges.Antoniadis, Olivier.Kraif, Claude.Ponton, Virginie.Zampa}@u-grenoble3.fr

[2] Université catholique de Louvain, Centre for English Corpus Linguistics
granger@lige.ucl.ac.be

**Summary**

In the first part of this article, we explore the background of computer-assisted learning from its beginnings in the early XIX[th] century and the first teaching machines, founded on theories of learning, at the start of the XX[th] century. With the arrival of the computer, it became possible to offer language learners different types of language activities such as comprehension tasks, simulations, etc. However, these have limits that cannot be overcome without some contribution from the field of natural language processing (NLP). In what follows, we examine the challenges faced and the issues raised by integrating NLP into CALL. We hope to demonstrate that the key to success in integrating NLP into CALL is to be found in multidisciplinary work between computer experts, linguists, language teachers, didacticians and NLP specialists.

## .1  Introduction

"Mechanised" learning is not a new idea; evidence of the first teaching machine goes back to 1809, in the USA. At that date, H. Chard patented a machine to teach reading called the *Mode of Teaching Reading* machine. However, it was another one hundred years before the first theories on these machines were expounded and the plans or prototypes of the first "theorized" machines were created. Thus, in 1912, (the psychologist) Edward Thorndike dreamed of a mechanical book, saying: "if, by some miracle of mechanical ingenuity, a book could be constructed in such a way that page 2 only became visible for those who have completed page 1 and so on, then much of that which currently requires personal instruction could be performed by the book" (Thorndike, 1913, quoted by Bruillard, 1997, p. 33-34). The theoretical foundations of mechanical programmed learning thus go back to the beginning of the XX[th] century in response to criticism of "traditional" teaching and so aimed primarily at permitting a rhythm of continuous learning which met the needs of the learner.

### .1.1  *The first teaching machines and programmed learning*

#### .1.1.1  *Pressey's machine*

The first "modern" teaching machine was developed by Sidney Pressey (1927) in 1924. It was a machine which could correct multiple choice questions with four buttons corresponding to the possible replies to the question put. The learner only moved on to the following question when he gave the right answer and the machine kept track of what the learner had done. Some, including Burrhus Frederic Skinner (1968), criticized Pressey for having based his machine on insufficient knowledge of the phenomenon of learning. These critics went on to focus their attention on these learning phenomena and to create programmed learning.

*.1.1.2   Programmed learning*

Skinner drew on the results of his work in behavioural psychology. From a theoretical standpoint, he held that conditioning operated to control learning mechanisms, which led him to imagine the creation of scientific teaching technology using linear programmed learning which could be administered by a teaching machine. For Skinner, effective learning was based on five principles:

- The active responding principle: the subject has to construct rather than simply recognize a correct response (multiple choice questions introduce errors that subjects would never make without it);
- the small steps principle: the material is broken down into small steps so that even the weaker learners can respond;
- the gradual progression principle;
- the self-pacing principle: learners should be able to set their own pace;
- the correct response principle: too many wrong responses discourage learners, so they must be guided.

Taking all this into account, the exercises on Skinner's machine, built in 1953, were on a serrated roller that the learner wound on himself. The questions appeared in a window, the learner entered his response in the white space reserved for this purpose, then compared it to the correct answer and activated a lever to move on to the following question. According to Skinner, no more than 10% of the learner's responses should be erroneous during one learning session.

But Skinner's programmed learning was criticised quickly and in 1959, Norman Crowder (1963) put forward an alternative system. His machine differs from Skinner's in a number of ways, in particular in the way the questions were introduced and the type of exercise. Crowder's machine first presented the learner with information which was followed by a multiple choice question. Unlike Skinner, who sought to limit errors, Crowder thought they had an important function. He considered that learning often involves distinguishing and discriminating between things. In his system, the learner's response was corrected and if it was right, he would move on to the next piece of information, but if he got it wrong, then he would be directed towards remedial exercises and then return to the exercise he had got wrong. If the learner answered several questions correctly, he would be given a short cut. This kind of teaching machine was thus better able to adapt to the individual learner and as such, had an important place in the development of programmed learning.

Maurice de Montmollin (1971) defined programmed learning as "a teaching method which makes it possible to transmit knowledge without the direct intermediary of a teacher or monitor, whilst taking into consideration the individual characteristics of each pupil ". This teaching is based on four principles: the principle of structuring the subject-matter (this involves selecting and presenting the subject-matter so as to facilitate comprehension and memorising), the adaptation principle (teaching has to be adapted to the pupil), the stimulation principle and the control principle. Progress is either linear (e.g. Skinner's machine) or branched (e.g. Crowder's machine).
Programmed learning opened up new avenues of research into methods and theories of teaching and learning and marked the beginning of computer-assisted instruction (CAI).

## .1.2  From CAI to CALL

Computer-assisted instruction (CAI) only really came into being at the beginning of the 1960s. At the start, the computer was only used to automate what had been done mechanically by the teaching machines. This involved primarily repeated exercises based on a behaviorist approach to learning and tutorials (which reinforced the learning) within a cognitive approach.

It was only in the 1970-80s, with work on expert systems and intelligent tutors, that the first attempts were made to make CAI "intelligent". This research was intended to make up for existing limitations, i.e.:

- to dialogue with learners in natural language;
- to select what needed to be taught;
- to anticipate, diagnose and understand the learner's errors;
- to improve teaching strategies and modify them in accordance with the learner.

It was during this period that the first micro-worlds came into being. Unlike the intelligent tutors whose aim was to turn computers into teachers and which were based on cognitive theories, the role of micro-worlds was to provide an environment for discovering abstract fields. The best known of these is Seymour Papert's Tortoise mini-language (1981). This was the first software to use a constructivist approach. Lastly, the 1990s saw the emergence of co-operative systems and interactive learning environments, founded on situated cognition theories. Although this software was not all used for teaching, it was nevertheless of relevance to CAI.

This period also saw the rapid rise of a new acronym: CALL (Computer-Aided Language Learning). Little by little, this new field defined its own issues and created its own specific tools. To our knowledge, this situation is a one-off; it has given rise to a significant number of specific studies and products, scientific journals and events where ongoing research is presented.

François Mangenot (1997) divides CALL exercises into five categories (cf. below). The importance of each category changes in accordance with the environment (DVD/Web), how it is being used (with a teacher, self-study, etc), the age of the learner, etc,

- **Comprehension tasks.** The rapid development of computers made it possible first to use audio sequences, then video sequences into the comprehension tasks. Most often, the related exercises are either in the form of multiple choice questions or objects (audio / video sequences, words, etc.) to click on or drag and the correction is given in terms of true/false, sometimes with some oral or written comments such as "bravo", "well done", etc.
- **Exercises intended to give learners discourse knowledge.** These generally involve jigsaws, sequencing and pairing. Correction is in terms of true/false.

- **Recordings of statements; oral exercises involving transformation of statements.** These exercises involve either repetition tasks or transforming statements. The learner can record and listen to himself. Some software uses the learner's voice to dub one of the characters, some let the learner look at his sonagram and compare it to that of the person whom he is imitating. In each case, auto correction is involved.
- **Simulations.** Mangenot identifies three types of simulations: those where the user can make the choices (as in those books "of which you are the hero") for example between going to the restaurant and going to visit a monument; those where the user has to familiarize himself with a task and replicate it, for example, create a CV; the third type of simulation involves linking images in order to make the system react. This type of exercise does not involve correction but comments are often provided.
- **Written productions.** Written production does not feature prominently in software. This is due to the fact that correction is too difficult. There are already problems with questions requiring necessarily short answers (e.g. cloze texts), but the problem is much more acute with free writing where syntax and semantics have to be taken into account. Consequently, the corrections given in CALL during writing exercises are either formulated in terms of standard corrections that the learner has to compare with his own writing or in the form of guidelines ("do not forget to say that…", "be more assertive", etc).

Aside from communication-based learning environments (between students, and with a teacher/tutor), CALL software contains very few self-corrective activities dealing with written or oral competence. The most widespread forms of exercises are undeniably those which are simplest for the computer to correct: generally, the learner is asked to choose one of a number of predetermined responses, as is the case for multiple choice, quizzes, sequencing exercises, drag and drop, pairing, gap exercises, etc. Despite the seeming variation in these activities, they all come down to a single paradigm, the closed question. CALL evolved at a time when it was limited by technological progress and certain 'mechanical' visions of the learning process and as a result still bears the mark of a narrow approach towards language didactics, with its ideas of linear progression, conditioned learning, where emphasis is placed on language structures rather than on communicative acts, a transmissive model of learning and on binary and summative evaluation. It is high time that thought is given to the full potential of CALL, making use of everything modern technology has to offer and avoiding the trap of adapting the practice to the tool rather than the reverse. Researchers in the field of Information and Communication Technology have already taken steps in this direction, but there is still a great deal which can be done through cross-fertilization with another field narrowly associated with information technology, the field of natural language processing (NLP).
.
.2 NLP comes to the aid of CALL

The first attempts at using NLP in language learning dates back to the 1980s, with a highpoint between 1985 and 1995. This view is upheld by Jung (2005**)** and also attested by a considerable number of symposia and European (Swartz et Yazdani, 1992), North American (Holland *et al.*, 1995) and French (Chanier *et al.*, 1993) publications. This period corresponds to the period during which the personal computer market expanded so rapidly and the first attempts at integrating ICT (Information and Communication Technology) into teaching mechanisms. During this period, a range of models and prototypes for analysis and language generation were brought out and it was these that served as a starting point for subsequent attempts at creating language learning systems. Records of this work can be found in all the journals dealing with language learning, such as *CALICO, Language Learning and Technology, ReCALL, CALL, ALSIC* and *System*.

Generally speaking, these attempts were initially rather crude and clumsy, and quickly reached the limit of their potential; large investment in human resources was necessary to get beyond the very restricted laboratory prototype stage. In addition, the issue of second language (not mother-tongue) learning versus acquisition was often not taken into account, which prevented any partnerships between computer experts/ CALL specialists and language and teaching specialists.

During the end of the 1990s, this problem was overcome, interdisciplinary meetings began to proliferate and joint work was carried out. Four new components were introduced into the information and communication system paradigm, with which the learner had to interact: one governing basic linguistic knowledge, another governing man/machine dialogue, a third the learner's knowledge and/or his errors and a fourth governing the system's didactic interventions. This paradigm has passed the test of time and CALL has become attractive to language and teaching specialists. The development of the Internet contributed to the implementation of the paradigm and learners can now easily access a rich variety of materials from a remote location while remaining in almost continuous contact with speakers, native speakers or otherwise, of the target language.

In this context, achievements are measured and progress is gradual. Every aspect of the CALL systems have been reexamined, redefined, challenged with the demands and realities of language learning. Current work aims to make use of all the "realistic" possibilities of results from NLP by adapting them to CALL objectives. Often, specific language processing programs are developed within a specific CALL application. In this context, NLP becomes more than a tool used in building CALL systems, it is enriched through the input of language teaching specialists on the issues at stake in language, its demands and limitations.

Although both written and oral forms of the language are involved, comparatively more work has been carried out on written language processing and this has had a direct impact on the number, and often quality, of the solutions and systems that have been created. Work and systems based on written language are definitely in the majority and they cover a more significant number of facets and situations of language learning.

Implicitly or explicitly, one idea underlies all this work: CALL requires multidisciplinary collaborative work, with all partners on an equal footing. Only if the issues arising from each of the disciplines/areas involved are tackled (Language Didactics, Information Technology (IT), Linguistics, and NLP) will it be possible to put forward solutions and operational systems, worthy of learners' interest and able to offer something extra over and above traditional systems and methods. For each partner, the essential task concerning each discipline or field lies in defining the part of the term "computer-assisted language learning " which concerns it; Thus, it is Didactics which is the most apt to define the term "learning", Linguistics the term "language", IT the term "computer-assisted". Each one of these definitions has to be put to the other partners, adapted to the constraints of the other partners' definitions, and contribute to the development of the overall solution.

## .3 Corpus processing for CALL

The use of corpora in language learning is a major field of application for NLP and, of course, for IT. IT has enabled language teachers to make use of the wealth of information and the range of language situations contained in corpora; it has given them a practically inexhaustible source of examples of "real" language, the type of language that the students will have to use in communicative situations. NLP makes it possible to seek not only character strings, but also the linguistic forms relating to lemmas, morphemes, morpho-syntactic features, functional relations, complex constructions and thus vastly extends the potential of corpus analysis.

Recent works have shown how to use corpora of authentic texts for activity generation (Selva 2002, Antoniadis *et al.* 2004). Corpora can be conceived as large repositories of examples that illustrate specific linguistic phenomena, which range from lexicon to morphology, syntax, phraseology, terminology and even translation (in the case multilingual corpus). These phenomena can easily be retrieved from tagged and lemmatised texts using simple parsing techniques, like Finite State Automata. Corpus mining as such may constitute an interesting resource for learners, in order to find occurrences of a given expression, to check a construction, to attest a meaning, etc. Deville *et al.* (2004) propose to use this functionality as an aid for comprehension.

Another interesting point is the possibility of searching new examples at each query (by a random selection of the parsed texts). By dynamically retrieving examples, new activities may be generated every time that the system is accessed: this is the case for Alfalex (Selva 2002), where gap-fill exercises allow practicing of French inflectional and derivational morphology, conjugations, prepositions, collocations, etc. with sentences that are extracted on-the-fly from a corpus. The data-driven learning approach has given rise to a large amount of work, resources and systems (Tribble and Barlow, 2001; Granger *et al.*, 2001; Granger, 2002), which could be greatly enhanced by the addition of simple NLP techniques.

By compiling and analyzing learner corpora, it also becomes possible to automate the analysis of the productions of the learners themselves. Learner corpora may result in two kinds of observations, useful to enhance error detection and analysis. First, they allow extraction of adapted mal-rules based on typical errors, i.e. errors that occur frequently in specific corpora (taking into account the learner proficiency and mother tongue). Second, learner corpus may be used as a benchmark to assess the results using various NLP techniques. As demonstrated by Metcalf & Meurers (2006), different types of word order errors call for different processing: those involving phrasal verbs (e.g. *they give up it*) can be handled successfully by means of instance-based regular expression matching while errors involving adverbs (e.g. *it brings rarely such connotations*) require more sophisticated parsing algorithms. A corpus containing learner errors may be used to to determine which errors fall in the scope of a specific technique and thereby design a more efficient error detection system.

From this perspective, we developed a tool, called eXXelant[1] (Granger *et al.*, forthcoming), a web-based error interface, designed for learner corpus browsing and concordancing. Taking as an input an XML formatted corpus, which contains error annotation and morphosyntactic tags, this tool allows extraction of examples using a query system that combines various kinds of criteria: error category, part-of-speech, corrected forms, error prone forms, learners' mother tongue and level. This tool has been first tested on a POS-tagged version of the FRIDA corpus[2]. Figure 1 illustrates the search for all past participle (i.e. "catégorie=verbe, trait=participe passé") errors (i.e. "erreur=oui") encountered in the corpus with the verb *avoir* (i.e. "lemme=avoir") as left-hand context. The results of the search are displayed in Figure 2.

---

[1] Exxelant stands for EXample eXtractor Engine for LAnguage Teaching. This tool has been developed in the framework of the Integrated Digital Language Learning (IDILL) Project, funded within the framework of the Kaleidoscope Network of Excellence.
http://www.noe-kaleidoscope.org/pub/

*Figure 1 : search for errors concerning the agreement of past partiple in "passé compose" tense.*

---

[2] The FRIDA learner corpus (FRench Interlanguage DAtabase) is a corpus of French as a Foreign Language compiled within the framework of the EU-funded FreeText project (Granger & al., 2001, Granger 2003).

*Figure 2 : Results of the previous search on the Frida-bis corpus*

Although *eXXelant* was not initially designed for direct use by learners, it has many features in common with Hegelheimer & Fisher's (2006) *iWRITE* system and could easily be adapted to perform similar activities of noticing and collaborative error solving. As pointed out by the authors, such material "can be used to raise learners' grammatical awareness, encourage learner autonomy, and help learners prepare for editing or peer editing."

The expansion of the Internet has made it possible to share and disseminate these resources and systems, which has probably been a strong contributing factor in the expansion of corpus use in language learning. Several CALL systems now use and exploit raw or annotated corpora; the care taken in compiling and annotating these corpora and their relevance to the specific situation is often a determining factor in the final quality of the system they are being used in.

## .4   Key issues for NLP - CALL integration

Aside from the issue of "whether one can do without NLP in CALL systems?" which has been and still is a concern of those working in the field (Chanier, 1998; Brown *et al.* , 2002, Antoniadis, 2004), there are four other key issues which need to be addressed:

- The contribution of NLP to CALL must be defined and evaluated. The potential contribution of NLP arises from the issues surrounding it and the aim it has set itself. Only NLP makes it possible to regard the language form not as a succession of signs with no meaning, but as elements of a two-tiered system (form and meaning). Within the language learning framework, the operation of each tier of this system has to be considered, made explicit, manipulated and put into practice in a targeted way; the links between the two tiers have to be demonstrated concretely and polysemy, a source of difficulty in language learning, should be given a place and dealt with appropriately. It is only currently possible to achieve this using NLP, to create systems and tools which will enable language teachers (and by extension students) to handle the language they are working with in the way it is actually structured, without the limitations imposed by basic IT techniques.

Like language teachers, a language learning system can only be considered valid, acceptable even, if it is able both to generate only correct language (not incorrect) and also to evaluate correctly students' language productions. In concrete terms, it is only possible to use NLP if the (often imperfect) results obtained using it (no NLP technique is 100% reliable) do not affect the integrity of the learning situation; ambiguity, a major problem in NLP, is only conceivable in CALL if it is dealt with through exchanges with the learner. For example, the automatic detection of an error should always be presented as a possibility, a presumption of error, if it is addressed to the learner. On the teacher's side, various support techniques can be envisaged: support in the generation of activities, in error detection, diagnostics, in analyzing the traces of the students' progress. It is in this sense that the contribution of NLP to CALL should be defined and evaluated. Current work is attempting, little by little, to determine the results of NLP which can be used effectively in building CALL systems.

- The better one knows a learner, the easier it is to define his or her learning needs. The individualized approach to learning is not a new idea, nor is it restricted to CALL. Although it is generally recognized that personalizing learning optimizes it, it is nevertheless indisputable that we are still waiting for systems capable of delivering this. It is not so much evaluating learners' knowledge that constitutes the major difficulty, but how to use this evaluation and, especially, how to automatically create activities relating to the learner's measured language level. NLP has something to contribute both in terms of evaluation tools and automatic activity creation. And while these tools are admittedly not a panacea, they nevertheless currently offer the only, albeit often partial, solution to these issues.

- One needs to be able to detect, explain and correct automatically learners' errors if the aim is to enable the learner to work autonomously. Like the issue of how to automatically generate explanations for the learners, the issue of how to auto correct students' productions "has been haunting" work in CALL since it came into being. A lot of work has been done on this, as can be seen from a recent issue of the review CALICO which is entirely devoted to this subject (Heift et Schulze, 2003). Nevertheless, going beyond a "true/false" type correction from a pre-established list of "truths" (or posting pre-established explanations) requires taking into account language features of answers given. These features then allow implanting the evaluation in the context of the teaching activity, in accordance with the learner's profile. Furthermore, instead of giving an evaluation in the form of a binary assessment (correct or incorrect), it may be more interesting to give potential elements of diagnosis, in order to help the learner to identify and correct the error by himself.

By defining the field, NLP can contribute a whole range of procedures and tools which can be used to address issues of evaluation and diagnosis through a constructive and cumulative approach. Nevertheless, this contribution of NLP should be considered within the context of its current potential to model and "measure" what is expected and reject error (the unexpected). As a result, analysis of errors[3] and generation of associated explanations cannot be directly implemented from NLP procedures taken from other applications, such as orthographical correction or traditional syntactic analysis; specific heuristics and methods still need to be invented.

- The tools and systems created should not require specific technical skills. As fields of knowledge, IT, just like NLP, Linguistics and Didactics, use their own, field-specific concepts and require the use of specific tools, generally founded on these concepts. For the outsider, the use of such tools can involve long periods of training and acquiring an understanding of a certain number of concepts from the field at the source of each tool. Within the framework of language learning, where teachers are *a priori* didactics specialists, their IT and NLP skills, even their linguistics skills, are generally comparatively much more limited and often virtually non-existent. This situation can make it problematic for teachers to famiIarize themselves with tools from fields outside their own scientific concern into their own field (and this is even more the case for students). Any CALL product intended *a priori* for language teachers should thus satisfy two imperatives: using it should require only a minimum of non-didactic skills and it should be able to handle didactic concepts, so that didactic or teaching solutions can be implemented.

More than twenty years after the beginning of work in "NLP and CALL", a number of prototypes and experimental systems have come into being (Chanier *et al.*, 1995; Brown *et al.*, 2002; Antoniadis *et al.*, 2004), and yet undeniably, commercial systems are extremely rare. This can only partially be explained by lack of research progress. In our opinion, there are two other main factors: ignorance of NLP on the part of language teachers[4], sometimes even on the part of computer experts, and the cost of resources and products coming out of NLP. As they are not standardised, they are difficult to use without modification and often require significant adaptations before being usefully deployed within CALL.

But there is no need for pessimism! A genuine effort at achieving some form of standardisation is at work in the NLP community, and an increasing number of more "generic" free software resources are coming out (concordancers, taggers, lemmatisers, etc.). Generally, these basic resources only require a minimum of investment in terms of adapting the input/output formats and setting the basic parameters. In the state of art, implementation of the simplest tools is likely to bring advantages which far outweigh this moderate investment. As for joint work between language teaching specialists and NLP specialists, current work in the field is tending to prove that a common working is in the process of being created (Antoniadis *et al.*, 2004; Brown *et al.*, 2002; Deville & Dumortier, 2004; Selva *et al.*, 2004).

## .5 Conclusion

The field covered by the application of NLP within CALL is broad. It covers numerous subjects some of which still need to be explicitly formalised, in particular as regards spoken language processing. As for written language processing, the numerous issues that have already been addressed have set just as many challenges, challenges which researchers are currently dealing with. Aid for error detection and diagnosis, generation and/or aid for activity generation, trace analysis, corpus excavation… all these avenues need to be explored, modelled (or remodelled), examined in the light of didactic objectives so that the tools which come out of the process are really useful for learners. One thing is sure however: if researchers working at the crossroads of NLP and CALL want to have a chance of seeing their work put into practice in language teaching, it is essential that they attach the utmost importance to the teaching capabilities of the product, a consideration which is undeniably still often marginal. The key lies in multi-disciplinary work and in sharing practices and techniques. Nevertheless, all the evidence suggests that the integration of NLP into CALL, which makes it possible to take into account the characteristics of the reference language and that of the learner, is happening in conjunction with the integration of principles of language didactics, the only way of ensuring that the learning objective is fully achieved by means of user-friendly and efficient CALL systems and tools. As language corpora constitute a common object of language teaching and NLP, they represent a natural ground to build integrated applications that meet the abovementioned requirements: in the very short term, we can expect many promising applications to emerge, especially in the field of learner corpus analysis and data-driven learning.

---

[3] It is not a question here of detecting only the existence of an error, but of calculating in addition the differential between the erroneous form and what should have been used in its place.

[4] The reverse being also true, joint work is often only a result of chance.